\documentclass[conference,10pt]{IEEEtran}
\IEEEoverridecommandlockouts

\usepackage{cite}
\usepackage[pdftex]{graphicx}
\usepackage[cmex10]{amsmath}
\usepackage{algorithmic}
\usepackage{array}
\usepackage{paralist}
\usepackage{amsfonts}
\usepackage{amsthm}
\usepackage{mathtools}
\usepackage{subcaption}
\usepackage{dsfont}
\usepackage[titlenumbered,ruled,vlined]{algorithm2e}
\newtheorem{theorem}{Theorem}

\makeatletter
\newcommand{\subalign}[1]{%
  \vcenter{%
    \Let@ \restore@math@cr \default@tag
    \baselineskip\fontdimen10 \scriptfont\tw@
    \advance\baselineskip\fontdimen12 \scriptfont\tw@
    \lineskip\thr@@\fontdimen8 \scriptfont\thr@@
    \lineskiplimit\lineskip
    \ialign{\hfil$\m@th\scriptstyle##$&$\m@th\scriptstyle{}##$\hfil\crcr
      #1\crcr
    }%
  }%
}
\makeatother

\begin{document}
\title{Bayesian Variational Federated Learning and Unlearning in Decentralized Networks}
\author{\IEEEauthorblockN{Jinu Gong}
\IEEEauthorblockA{\textit{School of Electrical Engineering} \\
\textit{KAIST}\\
Daejeon, South Korea}
\and
\IEEEauthorblockN{Osvaldo Simeone}
\IEEEauthorblockA{\textit{KCLIP Lab, CTR} \\
\textit{Dept Engineering, King's College London}\\
London, United Kingdom }
\and
\IEEEauthorblockN{Joonhyuk Kang}
\IEEEauthorblockA{\textit{School of Electrical Engineering} \\
\textit{KAIST}\\
Daejeon, South Korea}
}

\maketitle

\thispagestyle{plain}
\pagestyle{plain}

\begin{abstract}
Federated Bayesian learning offers a principled framework for the definition of collaborative training algorithms that are able to quantify epistemic uncertainty and to produce trustworthy decisions. Upon the completion of collaborative training, an agent may decide to exercise her legal ``right to be forgotten'', which calls for her contribution to the jointly trained model to be deleted and discarded. This paper studies federated learning and unlearning in a decentralized network within a Bayesian framework. It specifically develops federated variational inference (VI) solutions based on the decentralized solution of local free energy minimization problems within exponential-family models and on local gossip-driven communication. The proposed protocols are demonstrated to yield efficient unlearning mechanisms.
\end{abstract}


\section{Introduction}
\label{sec:intro}
\begin{figure}[t]
\centering
\includegraphics[width=5cm]{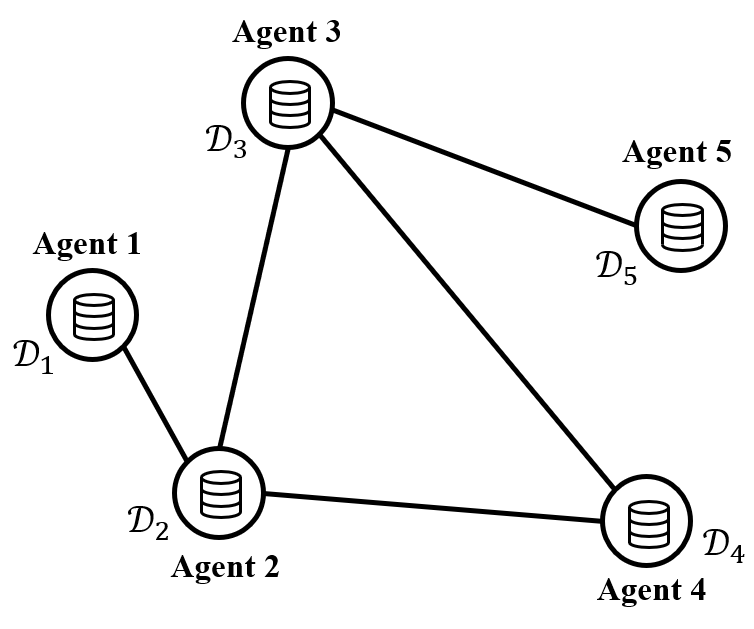}
\caption{Decentralized network with $K=5$ agents.}
\label{fig:main_network}
\end{figure}

Data produced and stored by different agents, ranging from individuals to companies and governments, can be useful to train joint machine learning models that outperform models trained separately on one of the data sets. To obviate the privacy cost, competitive loss, and communication load of a naive solution based on the sharing of data, federated learning was introduced as a rebranding of distributed training strategies that integrate local optimization steps at each agent and inter-agent exchange of model-centric, rather than data-centric, information \cite{li2020federated}. Upon the completion of collaborative training, an agent may decide to exercise her legal ``right to be forgotten'', which calls for her contribution to the jointly trained model to be deleted and discarded \cite{ginart2019making}. This paper studies federated learning and unlearning in a decentralized network (see Fig. \ref{fig:main_network}) within a Bayesian framework.

Federated learning is most often studied for a parameter server architecture, with a central server and multiple agents connected to it, within a frequentist framework \cite{li2020federated,kairouz2019advances}. Decentralized architectures, such as in Fig. \ref{fig:main_network}, describe a wider range of deployment use cases, from device-to-device (D2D) wireless systems \cite{xing2021federated} to inter-organizational networks, e.g., among banks or companies. Studies of decentralized training in such settings mostly adopt a frequentist formulation of the learning problem, whereby the goal is to collaboratively infer a single value for the model parameter vector. These works focus on convergence properties and empirical results for standard decentralized gradient methods \cite{xin2020decentralized}. Frequentist learning is limited in its ability to quantify epistemic uncertainty in the regime of limited data \cite{hullermeier2019aleatoric}, yielding potentially untrustworthy decisions \cite{guo2017calibration}. This motivates the study of Bayesian learning methods \cite{neal2012bayesian}.

In this work, we consider the setting in Fig. \ref{fig:main_network}, in which distributed agents communicate on an arbitrary D2D architecture with the goal of carrying out Bayesian learning or unlearning \cite{bui2018partitioned,nguyen2020variational}. Bayesian learning aims at collaboratively inferring a distribution in the model parameter space that approximates the global posterior distribution in the model parameter space given data at all agents. We develop variational inference (VI)-based solutions that generalize partitioned VI (PVI) \cite{bui2018partitioned,kassab2020federated} to decentralized networks under the assumption of a random walk schedule \cite{ayache2021private}. The approach recovers some of the solutions previously proposed in the context of distributed Bayesian signal processing \cite{djuric2012distributed}, and is also related to non-Bayesian social learning \cite{jadbabaie2012non}. The proposed approach, based on exponential-family parametrization, is shown to lend itself to efficient distributed unlearning mechanisms.

The rest of the paper is organized as follows. Sec. \ref{model} introduces system model and main definitions. Sec. \ref{learning} presents the proposed gossip-based federated variational learning protocol, which is extended in Sec. \ref{unlearning} to enable unlearning. Sec. \ref{experments} presents numerical results, and Sec. \ref{conclusion} concludes the paper.  
\section{System Model and Problem Definitions}
\label{model}

\subsection{Setting}
We consider a system with a set $\mathcal{K}=\{1,\ldots,K\}$ of $K$ agents connected by communication network illustrated by the undirected graph $\mathcal{G}=\{\mathcal{K},\mathcal{E}\}$ with corresponding adjacency matrix $A$, where $\mathcal{E}$ denotes the set of edges. The local data set $\mathcal{D}_{k}=\{z_{k,n} \}_{n=1}^{N_k}$ of agent $k\in\mathcal{K}$ contains $N_k$ data points, and the associated training loss for model parameter $\theta$ is defined as
\begin{align}
L_k(\theta)=\frac{1}{N_k}\sum_{n=1}^{N_k} \ell_k(z_{k,n}|\theta),
\end{align}
for some loss function $\ell_k(z|\theta)$. We also denote as $\mathcal{D}=\bigcup_{k=1}^K\mathcal{D}_k$ the global data set.

\subsection{Federated Bayesian Learning}
The agents collectively aim at obtaining the variational distribution $q(\theta)$ on the model parameter space that minimizes the \textit{global free energy} (see, e.g., \cite{kassab2020federated,bui2018partitioned})
\begin{align}
\min_{q(\theta)\in \mathcal{Q}} \bigg\{ F(q(\theta))=\sum_{k=1 }^K&\mathbb{E}_{\theta\sim q(\theta)}[L_k(\theta)]\nonumber\\
&+\alpha\cdot \mathbb{D}\big( q(\theta)\big\|p_0 (\theta)\big) \bigg\},\label{eq:global_fe}
\end{align}
where $\alpha > 0$ is a temperature parameter, $\mathbb{D}\left(\cdot\|\cdot\right)$ denotes Kullback–Leibler (KL) divergence, and $p_0(\theta)$ is a prior distribution. The variational posterior is constrained to lie in a set $\mathcal{Q}$ of distribution. When no constraints are imposed on set $\mathcal{Q}$, the optimal solution is given by the global generalized posterior distribution
\begin{align}
q^*(\theta|\mathcal{D})&=\frac{1}{Z}\cdot \tilde{q}^*(\theta|\mathcal{D})\label{eq:opt_sol_ori}\\ \textrm{where}\quad\tilde{q}^*(\theta|\mathcal{D})&=p_0 (\theta) \exp \left(-\frac{1}{\alpha}\sum_{k=1}^K L_k(\theta)\right),\label{eq:opt_sol}
\end{align}
which coincides with the conventional posterior $p\big(\theta|\mathcal{D}\big)$ when $\alpha=1$ and the loss function is given by the log-loss $\ell_k(z|\theta)=-\log p(z|\theta)$.
When the set $\mathcal{Q}$ of variational posteriors $q(\theta)$ is constrained, the solution of the problem (\ref{eq:global_fe}) is an approximation of the global generalized posterior (\ref{eq:opt_sol_ori})-(\ref{eq:opt_sol}). In federated Bayesian learning, the problem (\ref{eq:global_fe}) is addressed via a distributed training and communication protocol.

\subsection{Federated Bayesian Unlearning}
Assume that the system has obtained a, possibly suboptimal, solution  $q(\theta|\mathcal{D})\in\mathcal{Q}$ to problem (\ref{eq:global_fe}). In federated machine unlearning, we wish to remove from this distribution information about data set $\mathcal{D}_k \subset\mathcal{D}$ of some agent $k$. Ideally, we would do so by addressing problem (\ref{eq:global_fe}) from scratch using the data set $\mathcal{D}_{-k}=\mathcal{D}\setminus \mathcal{D}_k$ to obtain a variational posterior $q(\theta|\mathcal{D}_{-k})$, but this may be costly in terms of computation and convergence time. In federated Bayesian unlearning, the goal is devising decentralized protocols that are more efficient than training from scratch.

\subsection{Communication Protocol}
In this paper, we adopt a baseline gossip-based communication protocol based on a random walk on the graph \cite{ayache2021private}. Accordingly, at any time slot $i=1,2,\ldots,$ a node $k^{(i)}$ is scheduled to carry out local computing, and to (possibly) communicate with one of its neighbors when local computation is completed. Specifically, we adopt the Metropolis-Hastings (MH) scheduling \cite{ayache2021private}, whereby node $k^{(i)}$ chooses node $j$ uniformly at random from the set of neighbours $\mathcal{N}_{k^{(i)}}$. Then, for the given selected node $j$, node $k^{(i)}$ sets
\begin{equation}\label{eq:mh_rule}
k^{(i+1)}= 
\begin{dcases}
j& \text{w. p. }\min\left(1, \frac{\deg\left(k^{(i)}\right)}{\deg(j)} \right)\\
k^{(i)} & \text{otherwise}
\end{dcases}.
\end{equation}
Therefore, in the next time slot $i+1$, either the same node $k^{(i)}$ is scheduled, or a neighbor $j$ of node $k^{(i)}$ is scheduled.
 
\section{Gossip-Based Federated Bayesian Learning}
\label{learning}

In this section, we tackle the Bayesian variational learning problem (\ref{eq:global_fe}) over a set $\mathcal{Q}$ of variational distributions that, mimicking the optimal unconstrained solution (\ref{eq:opt_sol_ori})-(\ref{eq:opt_sol}), factorize as 
\begin{align}
q(\theta)=p_0 (\theta)\prod_{k=1}^K t_k(\theta),\label{eq:var_post_eq}
\end{align}
where the unnormalized distribution $t_k(\theta)$ is referred to as approximate local likelihood \cite{bui2018partitioned,kassab2020federated}. To optimize within the set $\mathcal{Q}$ defined by (\ref{eq:var_post_eq}), we follow a suboptimal local optimization procedure that proceeds coordinate-wise as in the expectation propagation and PVI frameworks \cite{bui2018partitioned,kassab2020federated}. At the end of each time slot $i$, the scheduled agent $k^{(i)}$ produces an updated variational posterior, which we refer to as $q^{(i)}(\theta)$. If $k^{(i+1)}\neq k^{(i)}$, this distribution is passed to agent $k^{(i+1)}$ before the beginning of time slot $i+1$.

\subsection{Gossip-Based Federated Variational Learning}
The proposed protocol, termed gossip-based federated variational learning (G-FVL), is defined as follows for any set $\mathcal{Q}$ satisfying (\ref{eq:var_post_eq}). Specific implementations based on the exponential family are detailed next.

\textbf{Initialization.} Randomly and uniformly choose initial agent $k^{(1)}$; set the current variational posterior distribution as $q^{(0)}(\theta)=p_0 (\theta)$; and $t_k^{(0)}(\theta)=1$ for all $k\in\mathcal{K}$.

\textbf{Step 1.} At the $i$-th iteration, agent $k^{(i)}$ updates the current global variational posterior $q^{(i-1)}(\theta)$ as a solution to the minimization of the \textit{local free energy}
\begin{align}
 \min_{q(\theta)\in\mathcal{Q}}\Bigg\{ F^{(i)}(q(\theta))=&\mathbb{E}_{\theta\sim q(\theta)}[L_{k^{(i)}}(\theta)]\nonumber\\
 &+\alpha\cdot \mathbb{D}\left(q(\theta)\Bigg\| \frac{q^{(i-1)}(\theta)}{t_{k^{(i)}}^{(i-1)}(\theta)}\right)\Bigg\}.\label{eq:loc_fe_min}
\end{align}

\textbf{Step 2.} Given the obtained solution $q^{(i)}(\theta)$, agent ${k^{(i)}}$ updates the local likelihood as
\begin{align}
t_{k^{(i)}}^{(i)}(\theta)&=\frac{q^{(i)}(\theta)}{q^{(i-1)}(\theta)}t_{k^{(i)}}^{(i-1)}(\theta),\label{eq:app_loc_like}
\end{align}
while non-scheduled agents $k'\neq {k^{(i)}}$ set $t_{k'}^{(i)}(\theta)=t_{k'}^{(i-1)}(\theta)$.

In a manner similar to \cite{kassab2020federated}, we can directly establish the following property of G-FVL. The proof is omitted since it follows the same steps as in \cite{kassab2020federated}.

\begin{theorem}\label{thm1}
The solution $q^*(\theta|\mathcal{D})$ in (\ref{eq:opt_sol_ori}) is the unique fixed point of G-FVL.
\end{theorem}

\subsection{Conjugate Exponential Family}
\label{learning_conj}
Problem (\ref{eq:loc_fe_min}) can be solved exactly in the special case in which the likelihood $p(z|\theta)$ defining the log-loss $\ell_k(z|\theta)=-\log p(z|\theta)$ and the prior $p_0 (\theta)$ form a conjugate pair in the exponential family (see, e.g., \cite{simeone2018brief}). This setting includes as a special case the beta-Bernoulli model studied in \cite{djuric2012distributed}. Accordingly, the likelihood is given as
\begin{align}
p(z|\theta)&=\exp\left(\eta_l(\theta)^T s_l(z)-A_l(\theta)+M_l(z) \right),\label{eq:conj_lh}
\end{align}
where $\eta_l(\theta)$ denotes the $D\times 1$ vector of natural parameters; $s_l(z)$ is the $D\times 1$ vector of sufficient statistics; $A_l (\theta)$ is the log-partition function; $M_l (z)$ is the log-base measure; and the prior is
\begin{align}
p_0(\theta)&=\exp\left(
\eta_0^T \begin{bmatrix}
\eta_l(\theta)
\\
-A_l(\theta)
\end{bmatrix}-A_0(\eta_0)+M_0 (\theta)\right)\nonumber\\
&=:\textrm{ExpFam}(\theta|\eta_0),\label{eq:conj_pri}
\end{align}
where $\eta_0$, $A_0(\eta_0)$, and $M_0(\theta)$ are the sufficient statistics, log-partition function, and log-base measure, respectively.
With these choices, the global posterior (\ref{eq:opt_sol_ori})-(\ref{eq:opt_sol}) is also in the same exponential family distribution as the prior (\ref{eq:conj_pri}), and is given as $p(\theta|\mathcal{D})=\textrm{ExpFam}\left(\theta|\eta_0+ \sum_{x\in\mathcal{D}}[s_l(x)^T,1]^T\right)$. Therefore, we can choose the approximate local likelihood without loss of optimality as $t_k(\theta)\propto \textrm{ExpFam}(\theta|\eta_k)$ for some natural parameter $\eta_k$.

With this selection, the optimal solution of problem (\ref{eq:loc_fe_min}) can be directly computed as (see, e.g., \cite{bui2018partitioned})
\begin{align}
q^{(i)}(\theta) &\propto \frac{q^{(i-1)}(\theta)}{t_{k^{(i)}}^{(i-1)}(\theta)}\exp\left(-L_{k^{(i)}}(\theta) \right)\label{eq:nat_para_sum_origin}\\
&\propto\textrm{ExpFam}(\theta|\eta^{(i)})\\
&\textrm{with}\quad \eta^{(i)}=\eta^{(i-1)}-\eta_{k^{(i)}}^{(i-1)}+\sum_{x\in\mathcal{D}_{k^{(i)}}}\begin{bmatrix}
s_l(x)
\\
1
\end{bmatrix},\label{eq:nat_para_sum}
\end{align}
where $t_{k^{(i)}}^{(i-1)}(\theta)=\textrm{ExpFam}(\theta|\eta_{k^{(i)}}^{(i-1)})$. Furthermore, the approximate local likelihood (\ref{eq:app_loc_like}) can be obtained as
\begin{align}
t_{k^{(i)}}^{(i)}(\theta)\propto\textrm{ExpFam}(\theta|\eta_{k^{(i)}}^{(i)}=\eta^{(i)}-\eta^{(i-1)}+\eta_{k^{(i)}}^{(i-1)}).\label{eq:loc_nat_update}
\end{align}
The resulting G-FVL algorithm is summarized in Algorithm \ref{alg:DVI_conj}. Note that agents only need to exchange the natural parameter $\eta^{(i)}$ of the variational posterior. 

By (\ref{eq:nat_para_sum})-(\ref{eq:loc_nat_update}), each agent $k$ simply adds $\sum_{x\in\mathcal{D}_{k}}[s_l(x)^T,1]^T$ to $\eta^{(i-1)}$ at the first iteration $i$ at which $k=k^{(i)}$; and sets $\eta^{(i')}=\eta^{(i'-1)}$ for any following iteration for which $k=k^{(i')}$. The process hence converges to the correct global posterior $p(\theta|\mathcal{D})$ after all agents have been visited once. By the properties of MH scheduling, for a fully connected network, this process requires on average $\sum_{i=1}^{K-1} (K-1)/(K-i)\approx (K-1)\log (K-1)$ iterations \cite{lovasz1993random}, which corresponds to the cover time of the random walk. More details on the cover time of general graphs can be found in \cite{lovasz1993random,aldous2002reversible}.
\begin{algorithm}
\caption{G-FVL for exponential family}
\label{alg:DVI_conj}
\SetAlgoLined
\SetKwInput{kwInit}{Initialization}
\kwInit{Set $i=1$; $\eta^{(0)}=\eta_0$; $\eta_k^{(0)}=0$ for all $k\in\mathcal{K}$; randomly and uniformly choose initial agent $k^{(1)}$;}
 \While{stopping criterion not satisfied}{Agent $k^{(i)}$ updates the global natural parameter $\eta^{(i)}$ via (\ref{eq:nat_para_sum}) or (\ref{eq:non_conj_update}), and the local natural parameter $\eta_{k^{(i)}}^{(i)}$ as in (\ref{eq:loc_nat_update})

The other agents $k\neq k^{(i)}$ set $\eta_k^{(i)}=\eta_k^{(i-1)}$
 
Agent $k^{(i)}$ chooses next node $k^{(i+1)}$ using (\ref{eq:mh_rule}) and shares $\eta^{(i)}$ with it

 $i\leftarrow i+1$
 }
\end{algorithm}
\vspace{-0.15cm}

\subsection{Non-Conjugate Exponential Family}
\label{learning_non_conj}
When the prior and the likelihood are not conjugate, the exact solution of problem (\ref{eq:loc_fe_min}) becomes intractable. For this more general case, as in \cite{kassab2020federated}, we assume that the approximate local likelihoods belong to an un-normalized exponential family distribution specified by natural parameter $\eta$, sufficient statistics $s(\theta)$, and log-based measure $M(\theta)$ as
\begin{align}
t_k(\theta|\eta_k)&=\exp\left(\eta_k^T s(\theta)+M(\theta)\right)\nonumber\\
&=\exp\left(\tilde{\eta}_k^T \tilde{s}(\theta)\right)\propto \textrm{ExpFam}(\theta|\tilde{\eta}_k), \label{eq:non_conj_app_lh}
\end{align}
with $\tilde{\eta}_k=[\eta_k^T, 1]^T$ and $\tilde{s}(\theta)=[s(\theta)^T, M(\theta)]^T$ being the augmented vectors of natural parameter and sufficient statistics, respectively. The prior distribution is also assumed to be in the exponential family with the same sufficient statistics as the approximate local likelihood (\ref{eq:non_conj_app_lh}), which we write as $p_0(\theta|\eta_0)=\textrm{ExpFam}(\theta|\tilde{\eta}_0)$. It follows that the variational posterior distribution (\ref{eq:var_post_eq}) can be expressed as
\begin{align}
q(\theta|\eta)&=\textrm{ExpFam}(\theta|\tilde{\eta})\quad \textrm{with}\,\,  \eta=\sum_{k=0}^K\eta_k \textrm{ and } \tilde{\eta}=\sum_{k=0}^K\tilde{\eta}_k, \label{eq:non_conj_post}
\end{align}
and $\tilde{\eta}=[\eta^T,1]^T$. In order to address problem (\ref{eq:loc_fe_min}), we follow natural gradient descent, yielding the local iteration $l$ \cite{khan2017conjugate}
\begin{align}
{\eta^{[l]}}\leftarrow{\eta^{[l-1]}}-&\rho\Bigg( {\eta_{k^{(i)}}^{[l-1]}}+\frac{1}{\alpha}\cdot\nabla_{\mu^{[l-1]}}\mathbb{E}_{ q(\theta|\eta^{[l-1]})}[L_{k^{(i)}}(\theta)]\Bigg),\label{eq:non_conj_update}
\end{align}
for local iteration index $l=1,\ldots,L$, with initialization $\eta^{[0]}=\eta^{(i-1)}$, output $\eta^{[L]}=\eta^{(i)}$, learning rate $\rho$, and with $\mu^{[l-1]}$ denoting the moment parameter corresponding to the natural parameter $\eta^{[l-1]}$ (see, e.g., \cite{simeone2018brief}). To estimate the gradient in (\ref{eq:non_conj_update}), we utilize the REINFORCE gradient as \cite{mohamed2020monte}
\begin{align}
&\nabla_{\mu}\mathbb{E}_{ q(\theta|\eta)}[L_{k^{(i)}}(\theta)]\nonumber\\
&\approx \frac{1}{S} \sum_{s=1}^S \left(L_{k^{(i)}}(\theta_s)-c\right)\cdot \nabla_{\mu}\log q(\theta_s|\eta),\label{eq:reinforce_grad}
\end{align}
where $\{\theta_s\}_{s=1}^S$ are i.i.d. samples drawn from distribution $q(\theta|\eta)$, $c$ is a baseline constant \cite{mohamed2020monte}, and
\begin{align}
\nabla_{\mu}\log q(\theta_s|\eta)&=\textrm{FIM}(\eta)^{-1}\nabla_{\eta}\log q(\theta_s|\eta)\\
&=\textrm{FIM}(\eta)^{-1}\left(s(\theta_s)-\mu\right),
\end{align}
where $\textrm{FIM}(\eta)$ denotes the Fisher information matrix (FIM) of the distribution $q(\theta|\eta)$. The overall algorithm is summarized in Algorithm \ref{alg:DVI_conj}. Unlike the conjugate case, the updates do not stop as soon as all agents have been visited.

\section{Gossip-Based Federated Bayesian Unlearning}
\label{unlearning}
In this section, we make the simple, but useful, observation that unlearning can be done efficiently in the settings considered in Sec. \ref{learning_conj} and Sec. \ref{learning_non_conj} in which the prior $p_0(\theta)$ and the approximate local likelihoods $t_k(\theta)$ belong to the same exponential family. As detailed next, this is due to the facts that (\emph{i}) each agent $k$ maintains an approximate local  likelihood  $t_k(\theta)\propto \textrm{ExpFam}(\theta|\eta_k)$ dependent on the natural parameter $\eta_k$; and (\emph{ii})  for any subset $\mathcal{K}'\subset \mathcal{K}$ of approximate local likelihoods in (\ref{eq:var_post_eq}), the resulting distribution can be computed as $q(\theta|\cup_{k\in\mathcal{K}'}\mathcal{D}_{k})=\textrm{ExpFam}(\theta|\eta_0+\sum_{k\in K'}\eta_k)$.

To elaborate, assume that the G-FVL algorithm discussed in the previous section is run for a certain number of iterations $I$, so that the current global variational parameter is $\eta^{(I)}$ and each agent $k$ has a local variational parameter $\eta_{k}^{(I)}$. Suppose now that we wish to delete data from agent $k$. Since the desired ``unlearned" posterior is
\begin{equation}
q(\theta|\mathcal{D}_{-k})= p_0(\theta)  \textrm{ExpFam}\Bigg(\theta\,\Bigg|\sum_{k'=1, k'\neq k}^K\eta_{k'}^{(I)}\Bigg),\label{eq:unlearn_new}
\end{equation}
one can easily obtain this result as follows. A scheduled agent $k^{(i)}\neq k$ forwards the current variational parameter $\eta^{(I)}$ to the next scheduled agent. At the first iteration that agent $k$ is scheduled, the agent $k$ computes $\eta^{(I)}-\eta_{k}^{(I)}$, which is forwarded to the next agent. This obtains exactly (\ref{eq:unlearn_new}). Note that the approach can be easily generalized to the forgetting of data from multiple devices.

For a complete graph, since the probability that we first reach to agent $k$ at $l$-th ($l \geq 2$) iteration is $(K-1)/K\times\left((K-2)/(K-1)\right)^{l-2}\times1/(K-1)$, the expected number of iterations for unlearning can be obtained as $
1/K+\sum_{l=2}^{\infty} l\times 1/K\times\left((K-2)/(K-1)\right)^{l-2}=1/K+K-1$. In contrast, following the discussion in Sec. \ref{learning_conj}, retraining from scratch excluding the $k$-th agent would take $\sum_{i=1}^{K-2} (K-2)/(K-1-i)\approx (K-2)\log (K-2)$ iterations on average \cite{lovasz1993random}.
\section{Experiments}
\label{experments}

\subsection{Beta-Bernoulli Model}
\begin{figure}[t]
   \centering
   \begin{subfigure}{0.12\textwidth}
       \includegraphics[width=\textwidth]{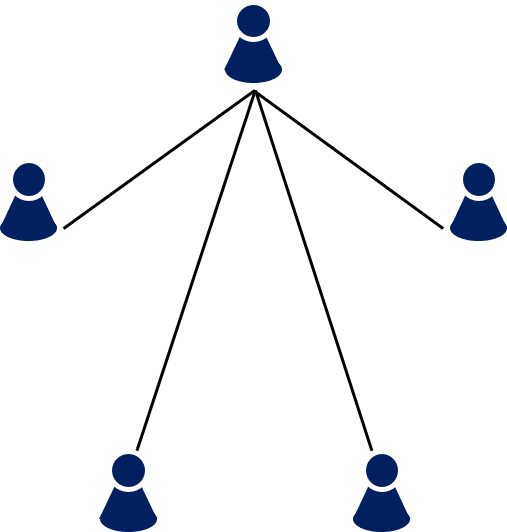}
       \caption{}
       \label{fig:table1}
   \end{subfigure}\hspace{0.03\textwidth}
   \begin{subfigure}{0.12\textwidth}
       \includegraphics[width=\textwidth]{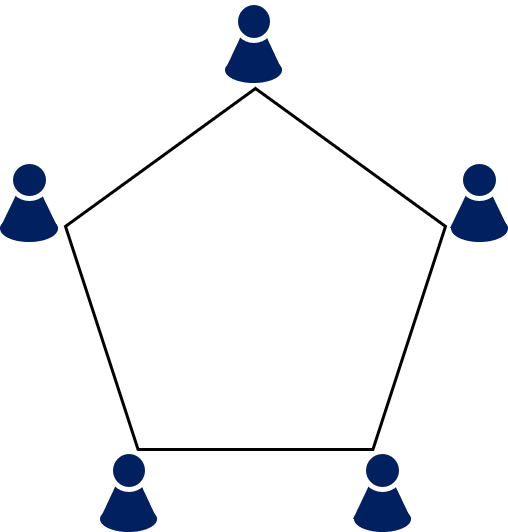}
       \caption{}
       \label{fig:table2}
   \end{subfigure}\hspace{0.03\textwidth}
   \begin{subfigure}{0.12\textwidth}
       \includegraphics[width=\textwidth]{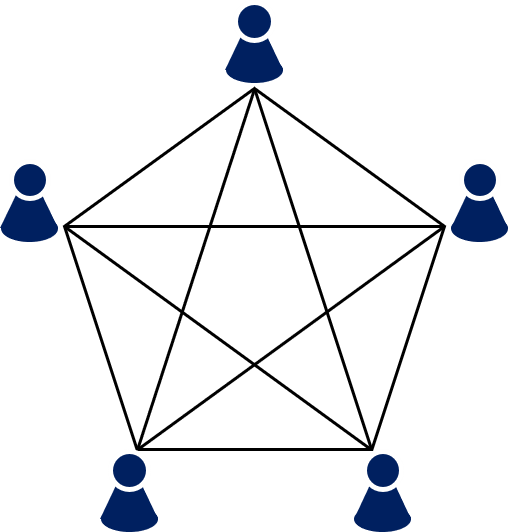}
       \caption{}
       \label{fig:table3}
    \end{subfigure}
    \caption{Examples of networks with $K=5$ agents. (a) star; (b) ring; (c) fully connected.}
    \label{fig:network}
\end{figure}

\begin{figure}[t]
\centering
\includegraphics[width=0.91
    \columnwidth]{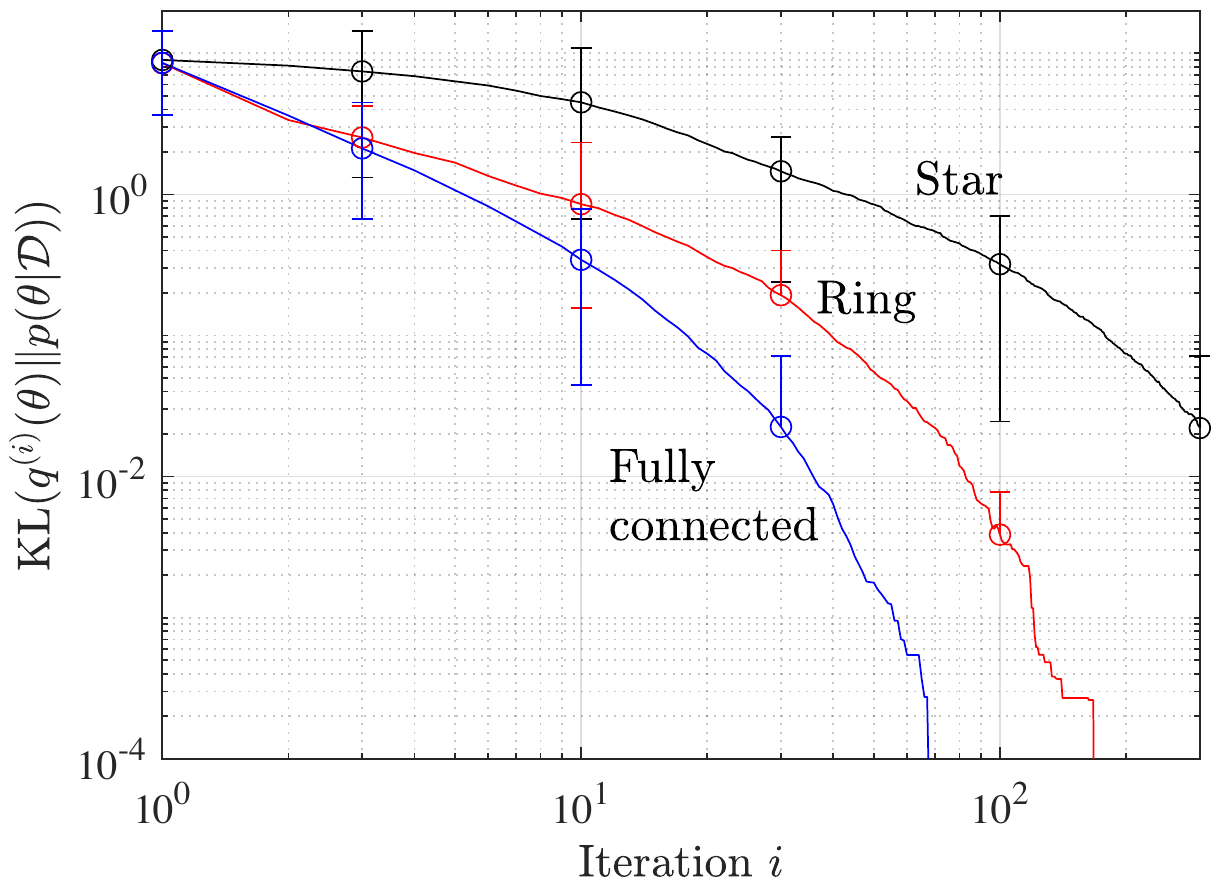}
\caption{KL divergence between the updated variational posterior and the true global posterior with $75\%$ credible intervals ($K=10$ agents under the three network architectures shown in Fig. \ref{fig:network}).}
\label{fig:betabern}
\end{figure}

For the first experiment, we focus on a standard conjugate exponential family pair consisting of the Bernoulli distribution $p(z|\theta)=\textrm{Bern}(z|\theta)$ and beta prior $p_0(\theta)=\textrm{Beta}(\theta|a,b)$. We consider three different network graphs, namely star, ring, and fully connected (see Fig. \ref{fig:network}) with $K=10$ agents, each having 100 data points. In Fig. \ref{fig:betabern}, we plot the average KL divergence between the current global variational posterior $q^{(i)}(\theta)$ and the true global posterior $p(\theta|\mathcal{D})$, with error bars representing $75\%$ credible intervals, for $a=b=2$ and $\alpha=1$. A fully connected network is seen to yield a faster convergence as compared to the standard star topology.

\subsection{Beta-Exponential Model}
\begin{figure}[t]
\centering
\includegraphics[width=0.91
    \columnwidth]{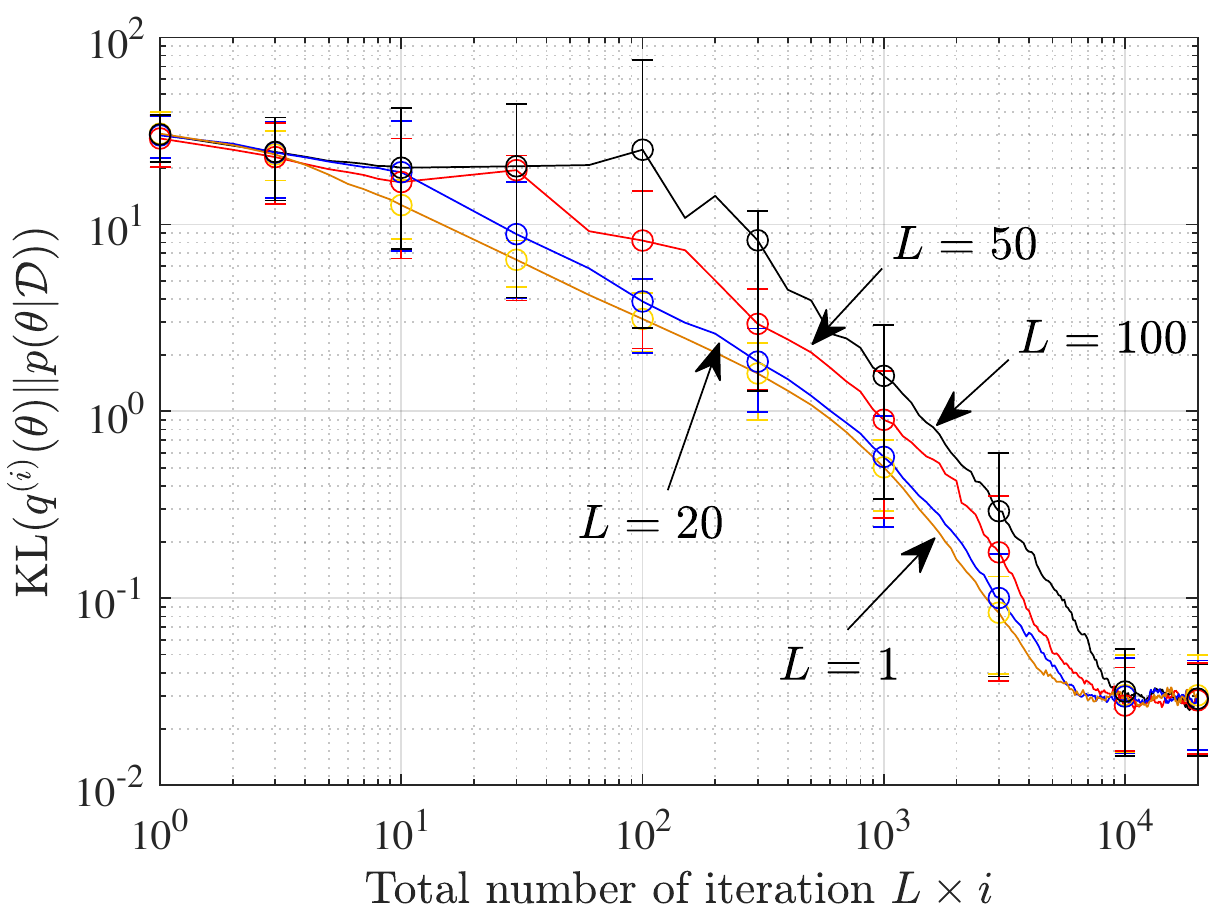}
\caption{KL divergence between the updated variational posterior using different values of local iteration $L$ and the true global posterior with $75\%$ credible intervals ($K=10$ agents, fully connected network).}
\label{fig:betaexp}
\end{figure}

We now study a non-conjugate model, in which the prior is the beta distribution $\textrm{Beta}(\theta|a,b)$ and the likelihood is
\begin{align}
p(z|\theta)&=\textrm{Exp}(z|\theta)\\
&=\frac{1}{\theta}\exp\left(-\frac{z}{\theta}\right)\mathds{1}(z\geq0).
\end{align}
We consider fully connected network with $K=10$ agents and apply G-FVL with parameters $\alpha=1$, $\rho_i=5\times10^{-3}$, $S=30$, and $c=0$. Fig. \ref{fig:betaexp} compares the KL divergence between the updated global posterior and the true global posterior using different values for the local iterations $L$. As can be seen, the convergence speed, measured in terms of the total number of local iteration $L\times i$, is degraded as the number $L$ of local iterations increases. However, this conclusion neglects the overhead required to communicate among agents, and it should be revisited if the performance is to be measured in terms of wall clock time.

\begin{figure}[t]
\centering
\includegraphics[width=0.91
    \columnwidth]{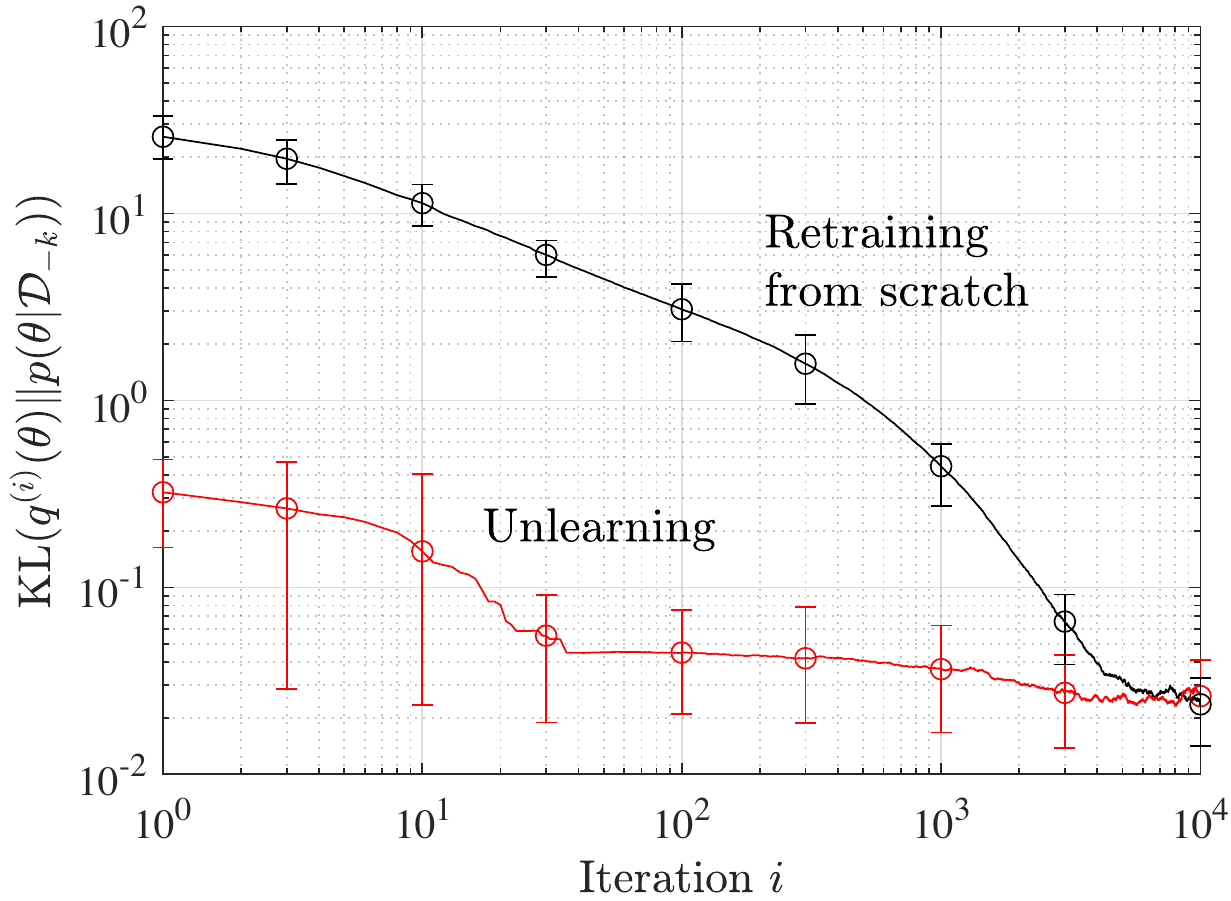}
\caption{KL divergence between the updated variational posterior via unlearning or retraining from scratch and the true global posterior with $75\%$ credible intervals ($K=10$ agents, fully connected network).}
\label{fig:kld_unlearning}
\end{figure}

In Fig. \ref{fig:kld_unlearning}, we plot the KL divergence $\textrm{KL}(q^{(i)}(\theta)||p(\theta|\mathcal{D}_{-k}))$ as a function of the iteration index $i$ for the unlearning setting in which we wish to delete the effect of data from agent $k=10$. The figure compares the simple approach proposed in Sec. \ref{unlearning} with the standard method that retrains from scratch. The latter scheme uses G-FVL by excluding agent $k=10$. The figure shows that as soon as agent $k=10$ is scheduled, which happens on average after $10$ iterations, the KL divergence is reduced to a value that is consistent with the asymptotic result obtained by the baseline training-from-scratch scheme.  
\section{Conclusion}
\label{conclusion}
This paper has studied Bayesian learning and unlearning within a variational inference framework by assuming distributions in the exponential family. Interesting generalizations include the derivation of non-parametric methods, for which the unlearning protocols cannot rely on the normalization property of the exponential family used here.

\section{Acknowledgments}
\label{sec:ack}
The work of J. Gong and J. Kang was supported in part by Institute for Information \& communications Technology Promotion (IITP) grant funded by the Korea government (MSIT) (No.2018-0-00831, A Study on Physical Layer Security for Heterogeneous Wireless Network), and in part by the MSIT (Ministry of Science and ICT), Korea, under the ITRC (Information Technology Research Center) support program (IITP-2020-0-01787) supervised by the IITP (Institute of Information \& Communications Technology Planning \& Evaluation). The work of O. Simeone was supported by the European Research Council (ERC) under the European Union's Horizon 2020 research and innovation programme (grant agreement No. 725731). 
\small

\bibliographystyle{IEEEtran}
\bibliography{ref}

\end{document}